\documentclass[11pt]{article}
\usepackage{coling2016}
\usepackage{times}
\usepackage{url}
\usepackage{latexsym}
\usepackage{graphicx}

\title{Truth Discovery with Memory Network}

\author{Luyang Li, Bing Qin, Wenjing Ren, Ting Liu \\
  \\
  Research Center for Social Computing and Information Retrieval,\\
  Harbin Institute of Technology,\\
  Harbin, China\\
  {\tt \{lyli, qinb,  wjren, tliu\}@ir.hit.edu.cn} \\}

\date{}

\begin{document}
\maketitle
\begin{abstract}
Truth discovery is to resolve conflicts and find the truth from multiple-source statements. Conventional methods mostly research based on the mutual effect between the reliability of sources and the credibility of statements, however, pay no attention to the mutual effect among the credibility of statements about the same object. We propose memory network based models to incorporate these two ideas to do the truth discovery. We use feedforward memory network and feedback memory network to learn the representation of the credibility of statements which are about the same object. Specially, we adopt memory mechanism to learn source reliability and use it through truth prediction. During learning models, we use multiple types of data (categorical data and continuous data) by assigning different weights automatically in the loss function based on their own effect on truth discovery prediction. The experiment results show that the memory network based models much outperform the state-of-the-art method and other baseline methods.

\end{abstract}

\section{Introduction}

In the age of information abundance, conflicts commonly exists in multiple-source statements about the same object. For example, different booking sites provide different boarding time about the same flight on the same date. The phenomenon seriously affects people's life. Truth discovery task is to find the most credible statement to resolve the confliction~\cite{Yin2008Truth,Dong2009Integrating,Galland2010Corroborating}. The development of the research benefit other natural language processing task like knowledge management~\cite{Poston2005Effective}, question answering~\cite{Banerjee2009Answer}, information retrieval~\cite{Olteanu2013Web} and so on.

Previous methods mostly take voting mechanism through predicting the truth~\cite{Yin2008Truth,Snow2008Cheap,Dong2009Integrating,Galland2010Corroborating,Dong2012Less,Wang2012On,Zhao2012A,Li2014A,Li2016A,Pasternack2010Knowing}. They consider the mutual effect between the reliability of sources and the credibility of statements; however, they mostly ignore the mutual effect among the credibility of statements. It is necessary to incorporate the effect from other statements of the same object when computing the credibility of the current statement. Another challenge is that there are multiple types of data in the real world, like categorical data and continuous data, which are both helpful in evaluating the reliability of sources. Few approaches jointly use multiple types of data to learn source reliability and predict the truth. CRH method first presente a framework to joint two types of data and is the state-of-the-art method. However it treats two types of data equally during the reliability evaluating which is intuitively not.

We propose memory network based models to learn source reliability and predict the credibility of statements.
Given an object, the statements from multiple sources with different reliability can be treated as context to each other. We use feedforward memory network (FFMN) and feedback memory network (FBMN) to incorporate the effect of the credibility between statements. The two models both use memory component to store and update the reliability of sources.
We propose a revised framework to utilize multiple types data and assign different weights based on the effect of different types data in model learning.
The proposed methods have been verified effectiveness through the experiments on public gold standard datasets. FFMN and FBMN outperform the state-of-the-art method.

We make the following contributions:

\begin{itemize}
\item We propose memory network based models to resolve truth discovery problem.
We convert the network-structure data into vectorial representation and input multiple-source statements of one object as context to predict the truth. The reliability of sources are stored as memories in FFMN and FBMN to help in computing the credibility of statements.

\item For better learning the reliability of sources, we proposed a framework to jointly utilize categorical data and continuous data. We weightily combine loss functions of two types of data, in which the weights are fine-tuned automatically through model optimization.

\item Through experiments, we verify that memory network based models outperform the state-of-the-art method, and specifically FFMN works better than other neural network models.
\end{itemize}

\section{Methodology}

We first make a formulation of the problem.
Then we introduce the framework and central functions.

\subsection{Problem Formulation}

In the real world, the \emph{statement} is a naturally existing information unit which is the description of a thing of an event. We can always draw an \emph{object} and a \emph{property} from the statement, which compose an \emph{entry}~\cite{Li2014Resolving}. The \emph{observation} is the \emph{value} of the \emph{entry}. According to the same entry, observations from different sources may contain conflicting values. The goal of the prediction model is to find the correct observation of the entry.

\fbox{\shortstack[l]{EXAMPLE 1. ``The flight AA-2446-LAX-DFW will take off at 01:19 PM" is a \emph{statement}. \\``The flight AA-2446-LAX-DFW" is an \emph{object} and ``takeoff time" is a \emph{property}.\\ ``The takeoff time of flight AA-2446-LAX-DFW" is an \emph{entry}.\\ The \emph{observation} of the \emph{entry} ``the takeoff time of flight AA-2446-LAX-DFW" is ``01:19 PM".}}

~\\
~~The \emph{object} may have more than one \emph{property} in a statement, and we separate it into multiple entries. In other words, we take an entry as the basic unit in the prediction model, and the credibility of the observations drew from same statement will be evaluated separately.

 \fbox{\shortstack[l]{EXAMPLE 2. ``The flight AA-2446-LAX-DFW will take off at 01:19 PM at gate 42B" has two \\ entries, ``the takeoff time of AA-2446-LAX-DFW"  and ``the gate of AA-2446-LAX-DFW".}}

~\\
~~Actually, there are two data types in the observations which are categorical data and continuous data. The categorical data is class-style data. The continuous data is a number, or can be converted into real number. By prediction model, we want to predict the correct category for categorical type data and predict the closest value to the true value for continuous type data.

 \fbox{\shortstack[l]{EXAMPLE 3. The ``gate 42B" is categorical data, and ``01:19 PM" can be treated as continuous \\ data by converting into minutes.}}

~\\
~~Suppose there are $N$ entries, each of which has $K$ observations offered by $K$ sources $\{s_1,s_2,...,s_K\}$. Given an entry $e_i$, the observations are the value set $V_i=\{v_{i1},v_{i2},...,v_{iK}\}$.

\subsection{Basic Framework}

In the direction of truth discovery, the lack of gold standard data is a challenge. CRH method~\cite{Li2014Resolving} uses an unsupervised framework to resolve the problem. Li et al. think reliable sources offer trustworthy observations which must be close to the truth. We cautiously think more trustworthy observations should be closer to the truth. Given an entry \emph{$e_i$}, \emph{D} is the function to compute the distance between the truth \emph{$t_i$} and the observation value \emph{$v_{ik}$}. Truth finding task is treated as an optimization problem. The optimization objective is to minimize the following loss function. The \emph{$r_{ik}$} stands for the credibility of the observation provided by source \emph{$S_k$}. We compute the credibility of the observation by using memory network based model, which is introduced in the Section 3.

\begin{equation}
f_{loss}=\sum_{i=1}^{N}\sum_{k=1}^{K}r_{ik}*D(t_i,v_{ik})
\end{equation}

The distance functions of categorical type data and continuous data are different. When entry $e_i$ belongs to the categorical data $U_{cate}$, the distance function $d_{cate}$ is as following.

\begin{equation}
d_{cate}(t_i,v_{ik})=
\left\{
\begin{array}{ll}
1, & v_{ik} \neq t_i \\
0, & v_{ik} = t_i
\end{array}
\right.
\end{equation}

If the observations of entry $e_i$ belong to continuous data $U_{con}$, the distance function $d_{con}$ is as following. The denominator of the function is the mean square error of value set $V_i=\{v_{i1},v_{i2},...,v_{ik},...v_{iK}\}$ according to the entry $e_i$. $\tilde{v_i}$ stands for the mean value of the value set $V_i$.

\begin{equation}
d_{con}(t_i,v_{ik})=\frac{|t_i-v_{ik}|} {\sqrt{(v_{i1}-\tilde{v_i})^2+(v_{i2}-\tilde{v_i})^2+...+(v_{iK}-\tilde{v_i})^2}}
\end{equation}

The truth $t_i$ which can minimize the overall weighted absolute distance is the weighted median~\cite{Li2014Resolving}. Given the observation set $\{v_{i1},v_{i2},...,v_{ik},...v_{iK}\}$ with the credibility set $\{r_{i1},r_{i2},...,r_{ik},...r_{iK}\}$, the weighted median of the set is $v_{im}$ which satisfies the following condition.

\begin{equation}
\sum_{k:v_{ik}<v_{im}}r_{ik}<\frac{1}{2}\sum_{k=1}^K r_{ik} ~~~\&~ \sum_{k:v_{ik}>v_{im}}r_{ik}\leq \frac{1}{2}\sum_{k=1}^K r_{ik}
\end{equation}

The loss function in our model is as following, which is the sum of the loss functions of categorical data and continuous data respectively. Specifically, the penalty values $\alpha$ and $\beta$ are learnt automatically through the model learning.

\begin{equation}
f_{loss}=\alpha\sum_{i \in U_{cate}}\sum_{k=1}^{K}r_{ik}*d_{cate}(t_i,v_{ik})+\beta\sum_{j \in U_{con}}\sum_{k=1}^{K}r_{jk}*d_{con}(t_j,v_{jk})
\end{equation}

Initialization. We use voting approach to compute the initial value of categorical data,and use averaging approach for continuous data. Although, the results would be not effected on condition that the optimization is convex~\cite{Li2014Resolving}.

\section{Memory Network based Model for Truth Discovery}

We would like to briefly introduce the memory network. Memory network is a framework with a long-term memory as inference component~\cite{Weston2014Memory,Sukhbaatar2015End,Tang2016Aspect}. It consists of a memory $m$ with four components $I$, $G$, $O$ and $R$.
The memory $m$ can be read and written to store the long-term information which is useful in prediction.
$I$ component learns the representation of inputs. $G$ component generates new memories based on a new input. $O$ component produces an output based on the new input and current memories. $R$ component converts the output into the response format.

In our problem, source reliability is long-term information to be used through predicting truth. Source reliability should be updated during learning a model by inputing each sample and be combined with input to generate a new output. We use memory $M=\{m_1, m_2,..., m_K\}$ to store the reliability of $K$ sources and updating the memories based on the input observations and current derivative of back propagation. The inputs are the vectorized observations and response of the model are the credibility  of observations. We learn the representation of the input data with memories by two types of models, feedforward neural network and feedback neural network. We call them Feedforward Memory Network (FFMN) and Feedback Memory Network (FBMN).

Pre-processing. The truth discovery data has network structure, and multiple observations of the same object have relationships.
We think embedding algorithms can learn the latent law of the structure and interaction among data.
We learn word embedding of data to vectorized every information from dataset. There are three kinds of data which are objects, properties and values. We use Word2Vec~\cite{Mikolov2013Efficient} to obtain the vector of each data. Specifically, we take a source and its observation of an entry (i.e., object, property and value) as a term of context in learning word embedding. We think the relevance between data can be learnt based on the context-based word embedding learning algorithm.

\subsection{Feedforward Memory Network}

\begin{figure*}[t]
\centering
\includegraphics[width=0.7\textwidth]{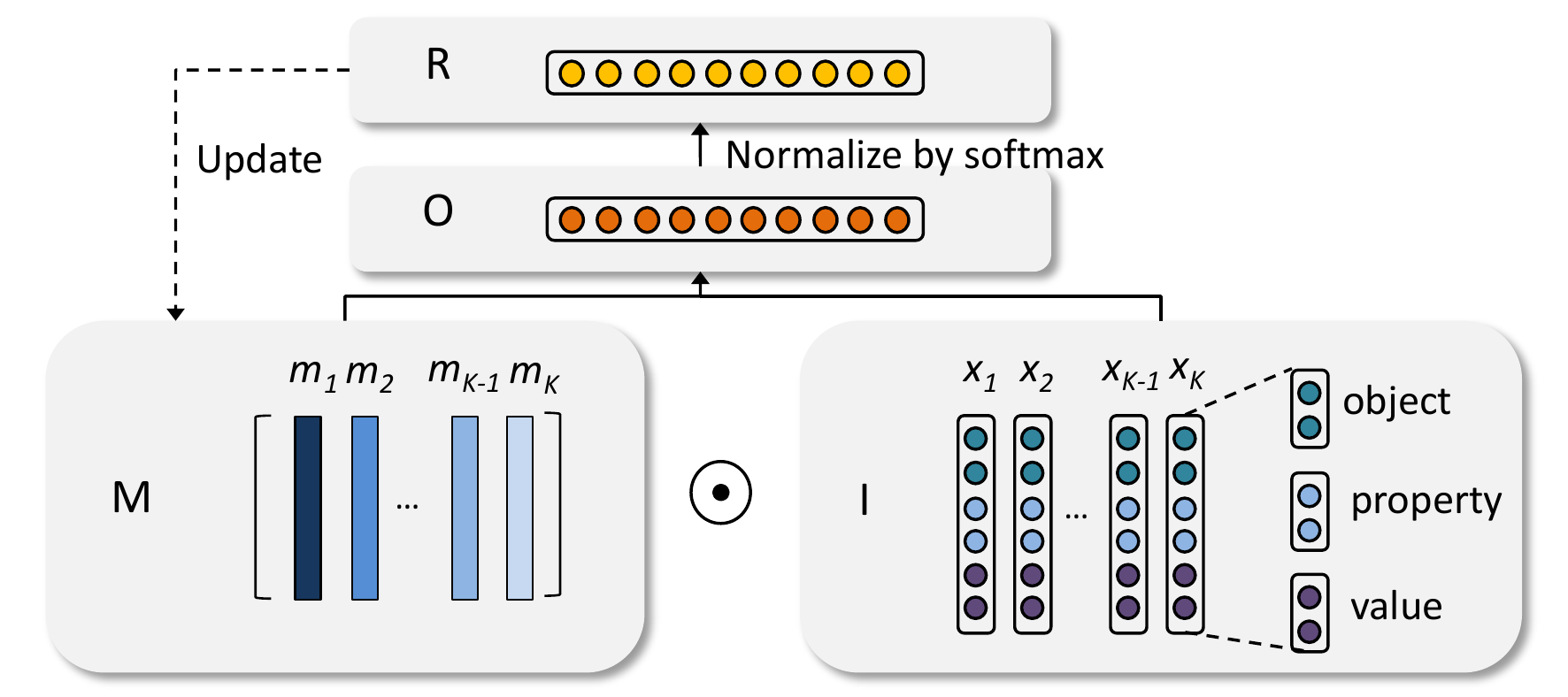}
\caption{The architecture of feedforward memory network.}
\label{figure:Feedforward memory network}
\end{figure*}

Feedforward neural network is like a directed acyclic graph which has no feedback through the network. Feedforward memory network is feedforward neural network with memory network mechanism. The architecture of FFMN is shown in Figure \ref{figure:Feedforward memory network}. The input $\{x_1, x_2,...,x_j,..., x_K\}$ of each iteration in $I$ component is the vectors of observations according to a same entry. The observation vector is the concatenation of the vectors of object data, property data and value data. $M$ stores $K$ memory vectors which stand for the reliability of sources. The memories and inputs are computed through element-wise multiplication. And $\sigma : R \rightarrow [0,1]$ is softmax function. The response of the FFMN is $\{r_1, r_2,...,r_j,..., r_K\}$ which is computed like in the following formula.

\begin{equation}
r_j=\sigma (m_j \odot x_j)
\end{equation}

\subsection{Feedback Memory Network}

Feedback neural network is a kind of network whose neurons feedback output to other neurons as input after a time step. Long short-term memory (LSTM) is a typical feedback neural network. LSTM uses hidden layer $H$ to store ``short term memory" and uses cell unit $C$ to store ``long term memory". It adopts input gate $x_t$ , forget gate $f_t$ to control the updating of ``long term memory" $c_t$, and adopts output gate $o_t$ to compute hidden vector at current timestep.

\begin{figure*}[t]
\centering
\includegraphics[width=0.99\textwidth]{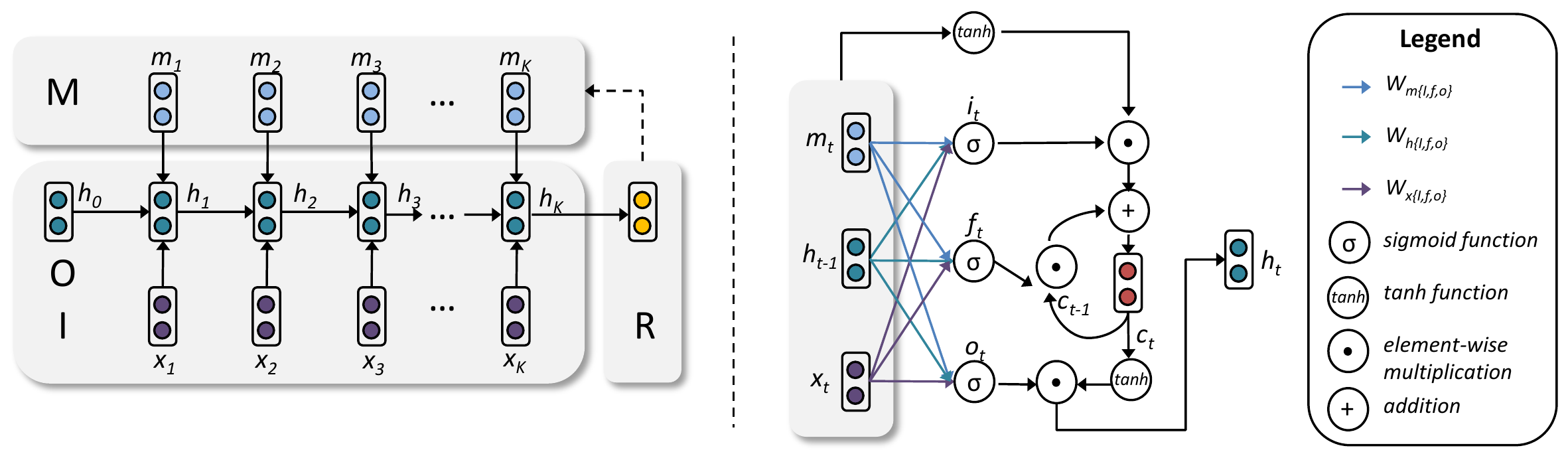}
\caption{The architecture (left) of feedback memory network and the detailed schematic (right) of a hidden vector generation.}
\label{figure:Feedback memory network}
\end{figure*}

Our feedback memory network (FBMN) is shown in Figure \ref{figure:Feedback memory network}. We add a memory component $M$ into LSTM to store the reliability of sources. The memory component $M$ plays a different role and also has a different operation by comparing with the cell $C$ in LSTM. The input of the model is a series of values from different sources according to a same entry. The time step is the order of input values. The response of the model is the result of softmax operation of the hidden vector of last time step. The updating of memories in $M$ is based on the response through the back-propagation of derivative.

The hidden vector $h_t$ is computed based on memory $m_t$, hidden vector on last time step $h_{t-1}$ and the current input $x_t$. Memory $m_t$ store reliability of the $t$th source. The formula are shown in the following, where $\sigma$ is the sigmoid function and $\odot$ is the element-wise multiplication. The detailed operating mechanism is shown on the right part of the Figure\ref{figure:Feedback memory network}.

\begin{equation}
i_t=\sigma (W_{ix}x_t+W_{ih}h_{t-1}+W_{ic}c_{t-1}+W_{im}m_t+b_i)
\end{equation}
\begin{equation}
f_t=\sigma (W_{fx}x_t+W_{fh}h_{t-1}+W_{fc}c_{t-1}+W_{fm}m_t+b_f)
\end{equation}
\begin{equation}
c_t=f_t\odot c_{t-1}+i_t\odot \tanh(W_{cx}x_t+W_{ch}h_{t-1}+W_{cm}m_t+b_c)
\end{equation}
\begin{equation}
o_t=\sigma (W_{ox}x_t+W_{oh}h_{t-1}+W_{oc}c_{t-1}+W_{om}m_t+b_o)
\end{equation}
\begin{equation}
h_t=o_t\odot \tanh (c_t)
\end{equation}

\section{Experiment and Analysis}

We present the experiment results to show the effectiveness of our memory network based models. We first introduce the dataset and evaluation criterions, and present the results and make an analysis. We also discuss the influence of parameters on the results.

\subsection{Datasets}

We use two public datasets~\cite{Li2012Truth} \footnote{The site of public datasets is http://lunadong.com/fusionDataSets.htm.} to demonstrate the effectiveness of the proposed method. The statistics of datasets are listed in Table \ref{table:Datasets}\footnote{We acquire datasets from the same site with Li et al. ~\cite{Li2014Resolving}, however, the statistics of datasets are a bit different.
Our statistics of datasets is done after data cleaning.}.
In view of existence of redundancy in datasets, which specifically causes multiple different values according to a same entry in the ground truth set, we eliminate the redundancy through data pre-processing.
The entries contained in ground truths are part of whole entries in the dataset. The ground truths are used only in the evaluation.

{\bf Stock Dataset}. Xian Li et al.~\cite{Li2012Truth} collected stock data from 55 sources on every work data in July 2011. There are 1000 stock symbols with 16 properties. The ground truths contain NASDAQ100 stocks and other 100 randomly selected stocks. These stock data are acquired by taking the majority of the values provided by five sources, nasdaq.com, yahoo finance, google finance, bloomberg and MSN finance. In order to verify the capacity of utilizing both categorical data and continuous data, Qi Li et al.~\cite{Li2014Resolving} treat the properties Volume, Shares Outstanding and Market Cap are considered as continuous type, and treat other properties as categorical type~\cite{Li2014Resolving}. We keep pace with them in the experiments.

{\bf Flight Dataset}. The flight dataset is collected from 38 sources over one month period (December 2011). It consists of 1200 flights with 6 properties. We treat departure gate and arrival gate as categorical type, and treat other properties as continuous type. We treat time data as real number by converting into minutes. The ground truths are about 100 randomly selected flights.

\begin{table}[h]
\begin{center}
\begin{tabular}[width=0.7\textwidth]{lll}
\hline
                   & Stock Data & Flight Data \\
\hline
Observations       & 12,056,684 & 2,703,448   \\
Entries            & 335,975    & 204,414     \\
Ground Truths      & 29,207     & 16,276      \\
\hline
\end{tabular}
\end{center}
\caption{Statistics of Datasets}
\label{table:Datasets}
\end{table}

\subsection{Evaluation Criteria}

We use two evaluation criteria, which are Error Rate and Mean Normalized Absolute Distance (MNAD)~\cite{Li2014Resolving}, to evaluate our methods. They are used to evaluate two types of data separately. The lower these two criteria are, the better the results are.

{\bf Error Rate}: Error rate is to compute the percentage of the wrong prediction of categorical data. If the output of the method is different from the ground truth, it is a wrong prediction. This evaluation criterion reflects the capability of methods to predict on categorical data.

{\bf MNAD}: MNAD is to evaluate the closeness between the prediction output and the ground truths on the continuous data. Because the values of different entries share different scales. The absolute distance between the prediction output $o_i$ and ground truth $t_i$ is normalized by the mean square error of the entry by given an entry $e_i$,.

\begin{equation}
MNAD=\frac{1}{M}\sum_{i=1}^{M}\frac{|t_i-o_i|} {\sqrt{(v_{i1}-\tilde{v_i})^2+(v_{i2}-\tilde{v_i})^2+...+(v_{iK}-\tilde{v_i})^2}}
\end{equation}

\subsection{Baseline Methods}

We use the following baseline methods to make a comparison with our methods.

    {\bf Mean}: Mean method averages all values of the same entry as the prediction which is used on continuous data.

    {\bf Median}:: Median method finds the median value of all values of the same entry as the prediction which is used on continuous data.

    {\bf GTM} ~\cite{ZhaoA}: Gaussian Truth Model (GTM) is a Bayesian probabilistic method which only work on continuous data.
        Note that this method only uses continuous data to learn the model and make a prediction, and insufficient data may lead to a lower performance in GTM compared with others.

    {\bf Voting}: The method takes the value with the highest occurrence as predicted value.

    {\bf Investment}~\cite{Pasternack2011Making}: In this approach, a source ¡°invests" its reliability uniformly on the observations it provides, and collects credits back from the credibility of those observations.

    {\bf PooledInvestment}~\cite{Pasternack2011Making}: PooledInvestment  linearly scale the credibility of observations which is different from Investment method.

    {\bf 2-Estimates}~\cite{Galland2010Corroborating}: This method is proposed based on the assumption that ¡°there is one and only one true value for each entry". If a source provides an observation for an entry, 2-Estimates assumes that this source votes against different observations on this entry.

    {\bf 3-Estimates}~\cite{Galland2010Corroborating}: 3-Estimates improves 2-Estimates by considering the difficulty of getting the truth for each entry, the estimation of which will affect the source¡¯s weight.

    {\bf TruthFinder} ~\cite{Yin2008Truth}: TruthFinder adopts Bayesian analysis, in which for each observation, its confidence is calculated as the product of its providers¡¯ reliability degrees. Similarity function is used to adjust the vote of a value by considering the influences between facts.

    {\bf AccuSim} ~\cite{Dong2009Integrating}: AccuSim also applies Bayesian analysis and it also adopts the usage of the similarity function. Meanwhile, it considers complement vote which is adopted by 2-Estimates and 3-Estimates.

    {\bf CRH} ~\cite{Li2014Resolving}: The method is the state-of-the-art method on the current datasets which take both categorical type and continuous type data into consideration through iteratively computing reliability of sources.

    {\bf Bi-LSTM}: Bidirectional LSTM is a variant of LSTM which is verified effective or even the state-of-the-art method in many NLP tasks.

\subsection{Truth Discovery Experiment}

We compare our FFMN and FBMN models with baseline methods. The experiment results in Table  ~\ref{table:truth discovery comparison} verify the effectiveness of our models.

\begin{table}[h]
\begin{center}
\begin{tabular}[width=0.9\textwidth]{lcccc}
\hline    &  \multicolumn{2}{c}{Stock Data} & \multicolumn{2}{c}{Flight Data}\\
Method & Error Rate & MNAD & Error Rate & MNAD \\ \hline
\emph{Previous non NN results} &    &    &   &   \\
Mean          & NA  & 7.1858  & NA  & 8.2894 \\
Median        & NA  & 3.9334  & NA  & 7.8471 \\
GTM           & NA  & 3.4253  & NA  & 7.6703 \\
Voting        & 0.0817  & NA &  0.0859 &  NA  \\
Investment    & 0.0983  & 2.8081 &  0.0919  & 6.4153 \\
PooledInvestment   & 0.0990  & 2.7940  & 0.0925  & 5.8562 \\
2-Estimates   & 0.0726  & 2.8509  & 0.0885  & 7.4347 \\
3-Estimates   & 0.0818 &  2.7749  & 0.0881 &  7.1983 \\
TruthFinder   & 0.1194 &  2.7140 &  0.0950  & 8.1351  \\
AccuSim       & 0.0726  & 2.8503  & 0.0881 &  7.3204   \\
CRH           & 0.0700 & 2.6445 & 0.0823 & 4.8613  \\
\hline
\emph{NN results} &    &    &   &   \\
LSTM          & 0.0884 & 2.4742 & 0.0013 & 1.8111  \\
Bi-LSTM          & 0.0737 & \textbf{1.4211} & 0.0170 & 1.7657  \\ \hline

\emph{Proposed method results} &    &    &   &   \\
FFMN          & \textbf{0.0207} & 1.5105 & \textbf{0.0008} & \textbf{1.2600}  \\
FBMN          & 0.0644 & \textbf{1.4211} & 0.0038 & 1.7711  \\
\hline
\end{tabular}
\end{center}
\caption{Classifier performance on cross-domain test data.}
\label{table:truth discovery comparison}
\end{table}

%

From the results, we can see our memory network based models outperform the previous methods. On stock data, FFMN has the best prediction capacity on categorical data which has a lowest error rate, and LSTM based models which are Bi-LSTM and FBMN perform better on continuous data which have lowest MNAD. On flight data, FFMN performs best both on categorical and continuous data. Among previous methods, CRH has the similar framework with our methods. The results between the two kinds of methods verify that the effectiveness of using neural network based model to resolve the truth discovery problem.

We make a comparison among neural network based methods and find memory network based models gain best results. FBMN has good MNAD results on continuous data. Specifically, FFMN makes an impression with a simple architecture, however, outperform other more complex models.

\subsection{Parameter Setting}

We have done a series of experiments to analyze the effect of parameters to the results of truth discovery. We make an analysis of the following parameters, embeddings learnt by different algorithms, different embedding length, different learning rates.

{\bf Embedding}: We adopt LINE~\cite{Tang2015LINE} word embedding learning algorithm to compare with Word2Vec~\cite{Mikolov2013Efficient} in truth discovery experiment. LINE can learn embedding from network-structure information which preserves both first-order proximity and second-order proximity. We use embedding in FFMN model and run experiments on flight data. The results show that embedding learnt by Word2Vec is more suitable for our problem. We also can see that LINE\_second order gains better than LINE\_first order.
\begin{table}[h]
\begin{center}
\begin{tabular}[width=0.7\textwidth]{lll}
\hline
                   & Error Rate & MNAD \\
\hline
word2vec           & \textbf{0.0008} & \textbf{1.2600}   \\
LINE\_first order   & 0.0017          & 1.7824     \\
LINE\_second order  & \textbf{0.0008}     & 1.9463      \\
\hline
\end{tabular}
\end{center}
\caption{The effect of different embeddings in truth discovery experiment.}
\label{table:Datasets}
\end{table}

{\bf Embedding Length}: We try multiple numbers of dimensionality from 50 to 300. We find the dimensionality has almost no effect on the results. Thus, the embedding length is set to 50 in final experiment.

{\bf Learning Rate}: We set the learning rate as 0.03, which is also verified with little impact on the results by trying from 0.0003 to 0.3.

\section{Related Work}

%

The truth discovery problem is to find the most credible statement. Most methods take a voting and similarity mechanism. More popular statements are more likely to be true. Similar statements have similar credibility. The statements from sources with similar reliability have similar credibility. The existing methods also take some important features into account, such as the reliability of sources, number of sources which post the same statements, difficulty of the statement, uncertainty in the information extraction~\cite{Pasternack2011Making}, similarity between statements and copying relationship, etc. Li et al. categorize the previous methods~\cite{Li2012Truth}.
The basic method is using voting strategy.
The web-link based methods share the similar strategy that the credibility of statements are computed based on the links between statements and sources. Corresponding methods are HUB~\cite{Kleinberg1999Authoritative}, AvgLog~\cite{Pasternack2010Knowing}, Investment~\cite{Pasternack2010Knowing} and PoolInvestment~\cite{Pasternack2010Knowing}. Specifically, Investment method takes the idea that the source ``invests" its reliability uniformly on its statements. The credibilities of statements are computed based on assessed reliability of sources. PoolInvestment method adds linear scaling on each entry through computing the credibility of statements.
The IR based methods are inspired by the similarity computing approach in information retrieval. Given a value of the object, the credibility is computed both based on the supporting sources and against sources. Corresponding methods are Cosine~\cite{Galland2010Corroborating}, 2-estimates~\cite{Galland2010Corroborating} and 3-estimates~\cite{Galland2010Corroborating}. Specifically, 3-estimates method take the likehood of correctness of voting on the value into account.

The Bayesian based methods apply Bayesian analysis, which are to predict the probability of a statement to be a truth based on observed information. The corresponding methods include TruthFinder~\cite{Yin2008Truth}, AccuPr~\cite{Dong2009Integrating}, AccuSim~\cite{Dong2009Integrating}, AccuFormat~\cite{Dong2009Integrating}, LCA~\cite{Pasternack2013Latent} and CRH~\cite{Li2014Resolving}. Specially, TruthFinder method considers similarity between statements, and AccuPr method consider that different statements on the same entry should be disjoint. LCA method is a probabilistic model which analyze latent credibility factors by using them as parameters to find the maximum a posteriori (MAP). CRH method~\cite{Li2014Resolving} use heterogeneous types of data which consists categorical data and continuous data to estimate the reliability of sources and predict truth.
The copying affected method discounts votes from copied observations through computing credibility, such as AccuCopy~\cite{Dong2009Integrating}.

%

\section{Conclusion}

Truth discovery is a fundamental research problem in natural language processing and data mining. Previous approaches mostly consider the mutual effect between the reliability of sources and the credibility of observations. The mutual effect among the credibility of observations to the same entry is also important, however, has not yet been appreciated. We use memory network based models to learn the latent relationship among observations of the same entry. Specifically, the proposed model adopts memory mechanism to learn the reliability of sources and incorporate it into the representing of observation credibility. We utilize multiple types of data and take their different contributions in the truth discovery by assigning weights automatically in the loss function.
The experiments show that our methods much outperform the state-of-the-art method on two public gold standard datasets. The feedforward memory network has the best performance.



\bibliographystyle{acl}
\bibliography{bibtex}

\end{document}